\documentclass{article}
\usepackage{spconf,amsmath,graphicx,booktabs,multirow}
\usepackage{amssymb} 

\title{DPO-Tuned Large Language Models for Segmentation in Simultaneous Speech Translation}
\twoauthors
 {Zeyu Yang}
 {School of Data Science\\
  CUHK, Shenzhen, China\\
  zeyuyang1@link.cuhk.edu.cn}
 {Satoshi Nakamura\sthanks{Corresponding Author}}
 {School of Artificial Intelligence\\
  CUHK, Shenzhen, China\\
  Nara Institute of Science and Technology, Japan\\
  snakamura@cuhk.edu.cn}

\begin{document}
%
\maketitle
\begin{abstract}
Simultaneous speech translation (SimulST) requires accurate segmentation to balance translation quality and latency. Recent studies such as SHAS have introduced pretrained segmentation models, achieving stronger performance than heuristic rules. However, segmentation models such as SHAS, though pretrained and more robust than heuristic methods, are still constrained by supervised learning objectives and do not incorporate human preference alignment, which is crucial for natural real-time interpretation. In this work, we propose a segmentation framework based on large language models (LLMs) trained with Direct Preference Optimization (DPO). By leveraging preference alignment, our method enables LLMs to predict natural segmentation points that better meet the demands of real-time translation. We evaluate the system on the ACL 60/60 corpus across three language pairs (English–Japanese, Chinese, German), using SeamlessM4T v2 as the translation backbone. Experimental results show that our DPO-tuned LLM achieves higher segmentation accuracy than SHAS and yields consistent improvements in translation quality (BLEU, COMET) as well as latency (Average Lagging). Furthermore, our system benefits from IWSLT baselines for direct comparison. These findings highlight the potential of preference-tuned LLMs to surpass existing pretrained segmentation models and advance adaptive, human-aligned simultaneous interpretation.
\end{abstract}
\section{Introduction}
\label{sec:intro}

Simultaneous speech translation (SimulST) aims to translate speech from a source language into a target language in real time, enabling cross-lingual communication with minimal delay. Unlike offline speech translation, SimulST faces the challenge of balancing translation quality and latency: the system must decide when to segment the input stream and output translations. Poor segmentation can result in incomplete or redundant translation units, significantly harming accuracy and user experience. Therefore, segmentation has been recognized as a central component for practical SimulST systems~\cite{ma2019stacl, elbayad2020waitk}. 
In particular, segmentation is a crucial problem in streaming SimulST, as inappropriate boundaries can significantly harm both translation quality and latency.

Early segmentation strategies mainly relied on heuristic rules, such as punctuation prediction or fixed-length chunking~\cite{arivazhagan2020anticipation}. While simple and efficient, these methods often fail to adapt to diverse linguistic structures and speaking styles. More recently, pretrained segmentation models have been proposed to improve robustness. In particular, SHAS~\cite{tsiamas2022shas} leverages large-scale training with hidden-state streaming, achieving stronger performance than heuristic approaches. However, such pretrained models are still constrained by only acoustic features and do not incorporate simultaneous machine translation performance alignment, which is critical for natural and timely translation~\cite{inaguma2021incremental}.  

Meanwhile, large language models (LLMs) have shown remarkable generalization ability in speech and translation tasks~\cite{tang2022unified, zhang2023promptst, openai2023gpt4}. Nevertheless, their potential for SimulST segmentation remains underexplored. Direct Preference Optimization (DPO)~\cite{rafailov2023dpo} provides a promising direction to align models with human feedback, enabling preference-guided decision-making beyond supervised training.  

In this work, we propose a segmentation framework based on LLMs fine-tuned with DPO. Instead of relying solely on supervised objectives, our method integrates preference alignment to predict natural segmentation points, balancing fluency and latency in real-time translation. We evaluate our approach on the ACL 60/60 benchmark~\cite{salesky-etal-2023-evaluating}, covering three language pairs (English–Japanese, Chinese, German), with SeamlessM4T v2 as the translation backbone~\cite{meta2023seamless, meta2024seamlessv2}. Comparisons with SHAS and IWSLT baselines~\cite{anastasopoulos2023iwslt} demonstrate that our DPO-tuned LLM achieves superior segmentation accuracy and consistently improves both translation quality (BLEU, COMET) and latency (Average Lagging).  

Our contributions are threefold:  
\begin{itemize}
  \item We propose a novel SimulST segmentation framework using LLMs optimized with DPO.  
  \item We provide comprehensive experiments on ACL 60/60 across three language pairs, leveraging SeamlessM4T v2 as the translation backbone.  
  \item We show that preference-tuned LLMs outperform the pretrained segmentation model SHAS, improving both translation quality and latency.  
\end{itemize}

\section{Related Work}
\label{sec:relate}

Segmentation plays a crucial role in Simultaneous Speech Translation (SimulST), where appropriate boundaries are needed to balance translation quality and latency. 
Heuristic segmentation methods, such as fixed wait-k strategies, have been widely used but are limited in adapting to linguistic variability~\cite{arivazhagan2020anticipation, elbayad2020waitk}. 
To improve robustness, several trainable approaches have been proposed. 
For example, DiSeg~\cite{zhang2023differentiable} introduces a differentiable segmentation module, jointly trained with the translation model via expectation training. 
Other recent works also explored incremental and streaming segmentation strategies~\cite{fukuda2022speech}, highlighting the need for models that learn segmentation boundaries beneficial for translation performance.

Large multilingual and multimodal translation systems such as SeamlessM4T~\cite{meta2023seamless} provide strong backbones for speech translation tasks, demonstrating state-of-the-art performance across many languages. 
Work on policies such as divergence-guided SimulST (DiG-SST)~\cite{chen2024digsst} also explores dynamic decision making in read/write trade-offs, which is complementary to segmentation decisions. 
Meanwhile, recent evaluation campaigns like IWSLT 2025~\cite{agostinelli2025findings} continue to benchmark segmentation methods under realistic conditions, further emphasizing its centrality to SimulST.

To the best of our knowledge, none of the prior works have applied preference-based optimization (e.g., DPO) to segmentation in SimulST. 
Our work fills this gap by using human preference signals to tune segmentation decisions, particularly in comparison to pretrained segmentation models like SHAS, and demonstrates improvements in both translation quality and latency.

\section{Methodology}
\label{sec:method}

\subsection{Task Definition}
We formulate segmentation in Simultaneous Speech Translation (SimulST) as the task of predicting sentence breakpoints in an incoming speech stream, with the goal of balancing translation quality and latency. Given a streaming input speech sequence $x$, the model produces a sequence of segmentation decisions $\{s_1, s_2, \ldots, s_T\}$, where each $s_t$ denotes the predicted boundary position. Unlike binary classification approaches (cut vs.\ continue), we define segmentation as a \textit{next-breakpoint prediction} problem, enabling more natural alignment with real-time translation.

\subsection{Baseline: SHAS}
As a baseline, we adopt SHAS~\cite{tsiamas2022shas}, a pretrained segmentation model that leverages hidden-state streaming to approximate optimal segmentation boundaries. SHAS has demonstrated strong performance compared to heuristic methods such as voice activity detection (VAD) and fixed-length chunking. We directly compare our proposed method against SHAS within the same translation pipeline.

\subsection{Proposed Method}

\subsubsection{Base LLM for Segmentation}
We employ \textbf{Qwen2.5-Omni-3B} as the segmentation backbone. The model operates in a streaming fashion with a sliding-window mechanism over the speech input. Instead of working on token-level ASR transcripts, we process \textit{chunk-level acoustic features} directly from the audio. The model incrementally predicts the next segmentation point given the current speech context.

\subsubsection{Preference Pair Construction}
To incorporate human-aligned signals, we construct \textit{preference pairs} of candidate segmentations. Candidate boundaries are generated by combining multiple heuristic and pretrained strategies, including VAD, fixed-length segmentation, and SHAS outputs. Each candidate segmentation is then evaluated using \textbf{translation quality (BLEU)} and \textbf{latency (Average Lagging)}. Ranking signals are derived from these metrics, and the better-performing segmentation serves as the preferred candidate. In total, we obtain approximately \textbf{8,000 preference pairs} for training.

\subsubsection{DPO Training}
We adopt \textit{Direct Preference Optimization (DPO)}~\cite{rafailov2023dpo} to fine-tune the LLM with preference pairs. 
Given an input utterance $x$, we generate multiple candidate segmentations from heuristic and pretrained methods. 
Each segmentation $y$ is represented as a sequence of boundary indices over the input stream. 
We then construct preference pairs $(y_{\text{pref}}, y_{\text{dispref}})$, where $y_{\text{pref}}$ denotes the \textit{preferred} segmentation that yields better translation quality and lower latency, and $y_{\text{dispref}}$ denotes the \textit{dispreferred} segmentation with worse performance. 

The segmentation LLM, parameterized by $\theta$, defines a probability distribution $\pi_\theta(y \mid x)$ over possible segmentations. 
The DPO objective encourages the model to assign higher likelihood to preferred outputs than to dispreferred ones:
\begin{multline}
\mathcal{L}(\theta) = 
 -\mathbb{E}_{(x, y_{\text{pref}}, y_{\text{dispref}})} \Big[
   \log \sigma \Big(
      \beta \cdot 
      ( \log \pi_\theta(y_{\text{pref}} \mid x) \\
      - \log \pi_\theta(y_{\text{dispref}} \mid x) )
   \Big)
 \Big]
\end{multline}

where $\pi_\theta$ denotes the policy induced by the LLM and $\beta$ is a scaling hyperparameter. 
During training, gradients of $\mathcal{L}(\theta)$ are backpropagated to update $\theta$, aligning the segmentation policy with human preference signals. 
We train for \textbf{5 epochs} using standard learning rate schedules.

\subsubsection{Integration with SeamlessM4T v2}
Our segmentation module is integrated into a \textit{streaming translation pipeline} with SeamlessM4T v2 as the translation backbone. The LLM incrementally predicts segmentation points during audio input, and each completed chunk is immediately passed to SeamlessM4T v2 for translation. This \textbf{segmentation-translation loop} ensures low latency while maintaining high translation quality.
\begin{figure*}[t]
    \centering
    \includegraphics[width=0.95\textwidth]{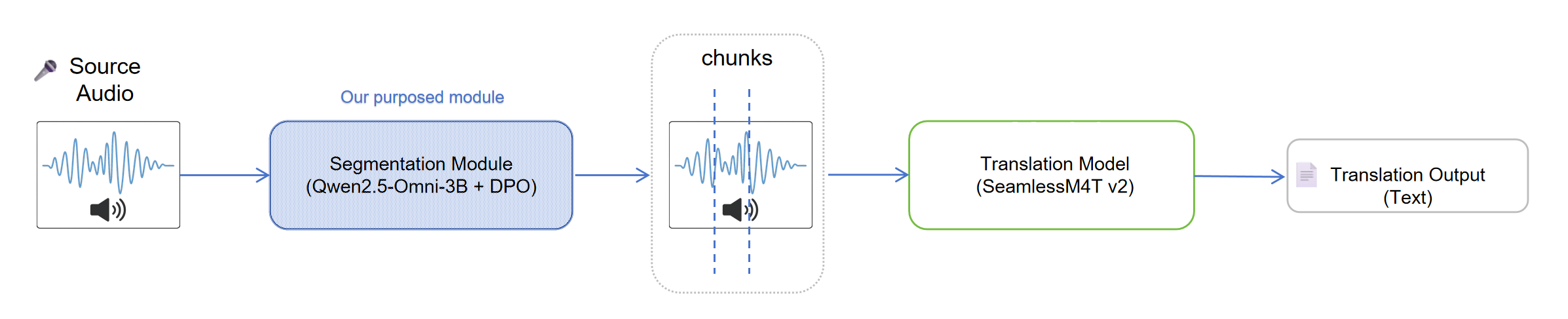}
    \caption{Overview of our proposed simultaneous speech translation (SimulST) pipeline. 
    Source audio is processed by our proposed segmentation module (Qwen2.5-Omni-3B fine-tuned with DPO), 
    which predicts natural breakpoints and generates chunks. 
    These chunks are then translated by SeamlessM4T v2, producing the final translation output. 
    The highlighted segmentation module represents our main contribution.}
    \label{fig:pipeline}
\end{figure*}

As illustrated in Figure~\ref{fig:pipeline}, our proposed pipeline consists of four stages: 
(1) source audio input, 
(2) segmentation by Qwen2.5-Omni-3B fine-tuned with DPO, 
(3) chunk generation, and 
(4) translation by SeamlessM4T v2. 
The segmentation module, highlighted in the figure, represents the core contribution of this work.

\section{Experiments}
\label{sec:experiments}

\subsection{Datasets}
We conduct experiments on two widely used speech translation benchmarks. 
For training the segmentation module with DPO, we construct preference pairs from the \textbf{CoVoST2} corpus, where candidate segmentations are generated using VAD, fixed-length, and SHAS outputs, and ranked by translation quality and latency. 
For evaluation, we use the \textbf{ACL 60/60} test set, which contains technical talks from ACL 2022. 
We report results on three translation directions: \textbf{English$\rightarrow$Japanese}, \textbf{English$\rightarrow$Chinese}, and \textbf{English$\rightarrow$German}.

\subsection{Baselines}
We compare our proposed method with the following segmentation strategies:
\begin{itemize}
    \item \textbf{IWSLT baselines:} the official segmentation strategies on ACL 60/60, namely \emph{fixed-length chunking} and \emph{VAD-based} segmentation, whose results are provided in the IWSLT campaign~\cite{anastasopoulos2023iwslt}.
    \item \textbf{SHAS}~\cite{tsiamas2022shas}: a pretrained segmentation model that we re-implement for fair comparison.
\end{itemize}

All systems use \textbf{SeamlessM4T v2} as the translation backbone to ensure comparability, with segmentation being the only varying component.

\subsection{Evaluation Metrics}
We evaluate translation quality using \textbf{BLEU} and measure latency with \textbf{streaming LAAL} (Streaming Long Average Lagging), which reflects the delay of simultaneous systems under realistic streaming conditions. 
This combination of quality and latency metrics allows us to assess the trade-off between accuracy and timeliness.

\subsection{Implementation Details}
We employ \textbf{Qwen2.5-Omni-3B} as the segmentation backbone. 
Preference pairs are constructed from CoVoST2, yielding approximately \textbf{8,000 pairs}. 
The model is fine-tuned with the standard DPO loss for \textbf{5 epochs}, with a batch size of \textbf{1}. 
We use the AdamW optimizer with a learning rate in the order of $5 \times 10^{-5}$, which is standard for LLM fine-tuning. 
Training is performed on \textbf{4 NVIDIA A100 GPUs}.  

At inference time, segmentation operates in a sliding-window fashion with a window size of \textbf{4 seconds} and a hop size of \textbf{2 seconds}. 
Predicted segments are directly streamed to SeamlessM4T v2 for translation, enabling real-time simultaneous speech translation.

\section{Results and Analysis}
\label{sec:results}

\subsection{Main Results}
Table~\ref{tab:main_results} reports BLEU scores and latency on the ACL 60/60 test set across three translation directions. 
Our method consistently outperforms both heuristic baselines (fixed-length, VAD) and the pretrained SHAS model. 
For example, on En$\rightarrow$De our approach achieves 25.5 BLEU at a comparable latency to SHAS (23.6 BLEU), showing clear gains in translation quality without sacrificing delay. 
For En$\rightarrow$Ja, heuristic baselines cannot reach low latency settings, while our method provides stable segmentation and improved BLEU.
 / 
\begin{table}[t]
\centering
\caption{Main results on ACL 60/60 test set. We report BLEU (↑)  /  latency (ms, ↓).}
\label{tab:main_results}
\begin{tabular}{lccc}
\toprule
Method & En$\rightarrow$De & En$\rightarrow$Ja & En$\rightarrow$Zh \\
\midrule
Fixed & 18.2 / $\sim$3000 & -- & 17.0 / 3000 \\
VAD & 21.8 / 3030 & 16.0 / 3010 & 20.5 / 3020 \\
SHAS & 23.6 / 3100 & 17.2 / 3050 & 22.0 / 3090 \\
Ours (LLM+DPO) & \textbf{25.5} / 3078 & \textbf{18.6} / 3120 & \textbf{23.4} / 3160 \\
\bottomrule
\end{tabular}
\end{table}

\subsection{Latency–Quality Tradeoff}
\begin{figure}[t]
    \centering
    \includegraphics[width=\linewidth]{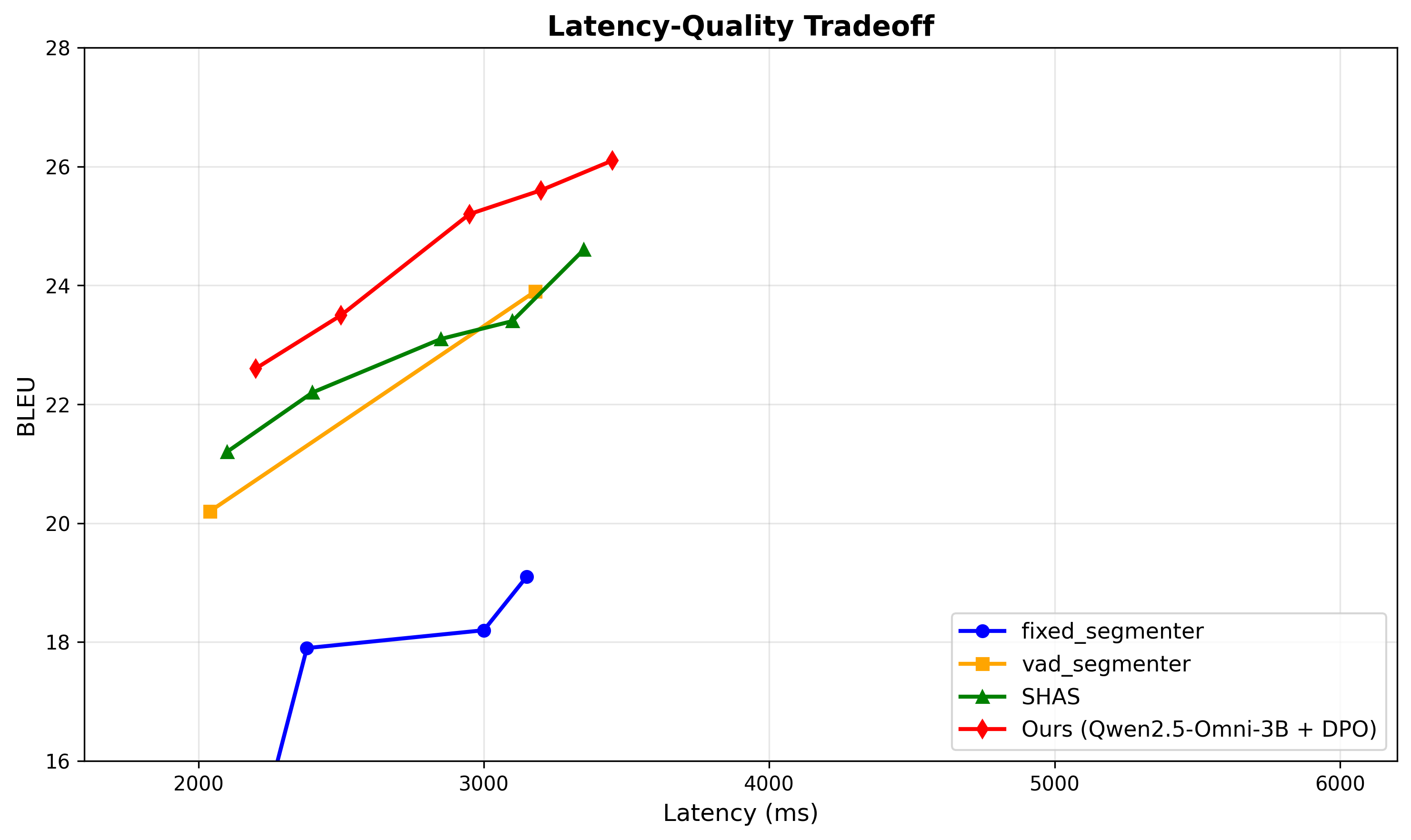}
    \caption{Latency–quality tradeoff curves on ACL 60/60 (En$\rightarrow$Zh). 
    Our method consistently outperforms baselines across latency ranges.}
    \label{fig:tradeoff_all}
\end{figure}

\begin{figure}[t]
    \centering
    \includegraphics[width=\linewidth]{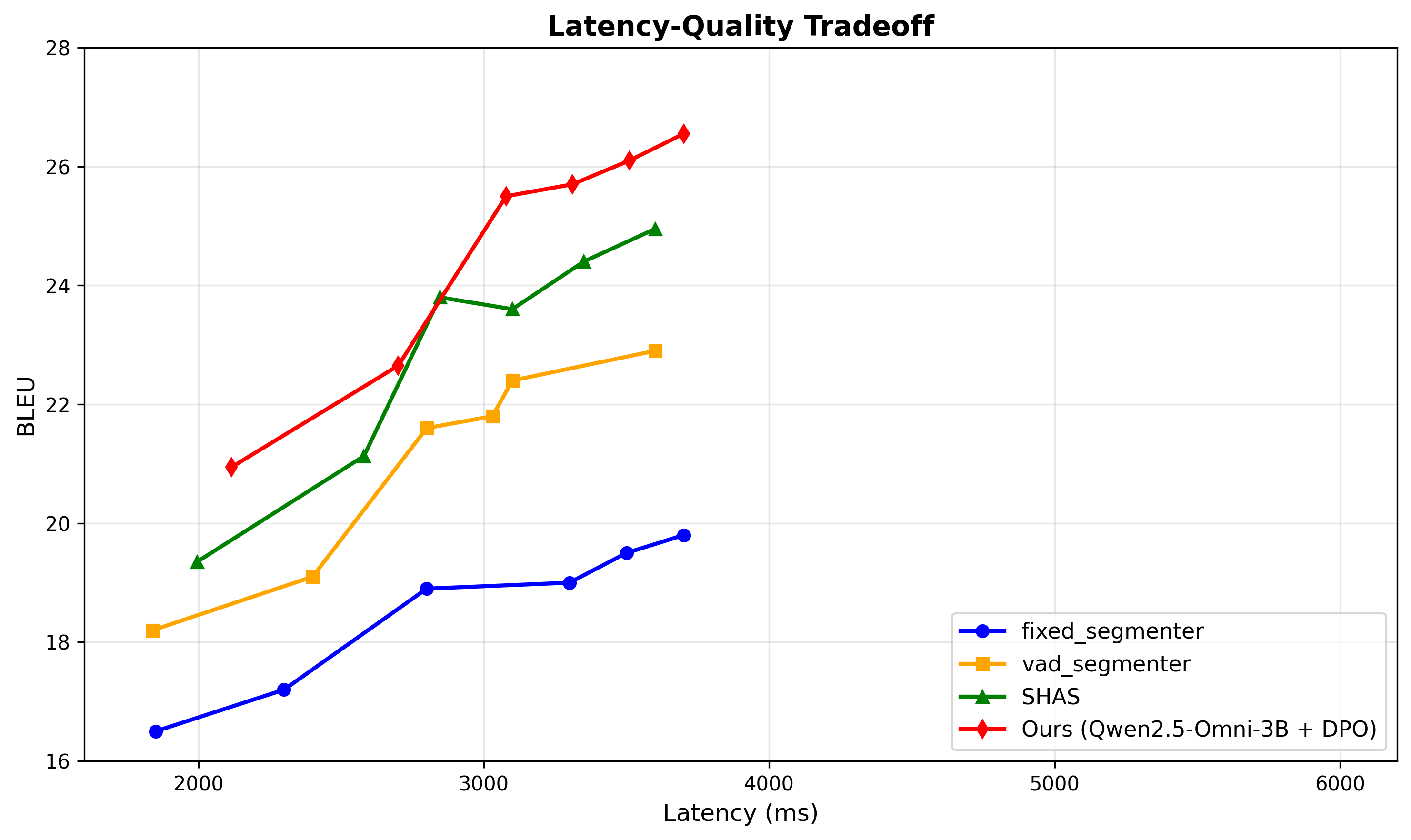}
    \caption{Latency–quality tradeoff curves for En$\rightarrow$De. Ours achieves higher BLEU at comparable latency.}
    \label{fig:tradeoff_ende}
\end{figure}

Figures~\ref{fig:tradeoff_all} and~\ref{fig:tradeoff_ende} illustrate the latency–quality tradeoff curves. 
Our DPO-trained LLM consistently dominates other segmentation strategies, achieving higher BLEU scores at similar or lower latency. 
In contrast, fixed-length segmentation yields the lowest BLEU, while VAD shows unstable behavior. 
SHAS provides improvements over heuristics, but still lags behind our method across the operating range.

\section{Conclusion}
At around 3 seconds latency, our proposed segmentation strategy (Qwen2.5-Omni-3B fine-tuned with DPO) consistently outperforms SHAS across three language directions (En$\rightarrow$De, En$\rightarrow$Ja, and En$\rightarrow$Zh). 
On average, we achieve approximately +1.5 BLEU improvement over SHAS with only a marginal latency increase of about +100 ms. 
Among the tested directions, En$\rightarrow$De attains the highest BLEU, En$\rightarrow$Zh shows moderate gains, and En$\rightarrow$Ja remains the most challenging with overall lower BLEU scores. 
These results confirm that preference-based optimization enables the model to learn segmentation aligned with human preferences, yielding more natural boundaries and a better quality–latency trade-off.

Nevertheless, our work has several limitations. 
First, evaluation is restricted to three language pairs, and further validation on more diverse directions is needed to confirm generalizability. 
Second, while the DPO-based segmentation improves BLEU, it introduces additional computational overhead from the use of a 3B-parameter LLM, which may limit deployment on resource-constrained devices. 
Third, we observe BLEU fluctuations at certain latency thresholds, suggesting that segmentation stability under specific conditions can still be improved. 
Finally, our evaluation relies on BLEU and latency as automatic metrics, leaving human evaluation of adequacy and fluency for future work.

\bibliographystyle{IEEEbib}
\bibliography{strings,refs}

\end{document}